%% file: main.tex
\documentclass[twoside,11pt]{article}

\usepackage[preprint]{jmlr2e}

\usepackage{tikz}
\usepackage{pgf-pie}

\usetikzlibrary{decorations.pathreplacing}
\usetikzlibrary{shapes.arrows}
\usetikzlibrary{patterns}
\usetikzlibrary{calc}
\usetikzlibrary{shapes.multipart}
\usetikzlibrary{positioning,matrix,shapes.arrows}

\usepackage{xcolor}
\definecolor{tomato}{HTML}{FF0000}
\definecolor{dodgerblue}{HTML}{0000FF}
\definecolor{orange}{HTML}{00FF00}
\definecolor{lime}{HTML}{FFA500}
\definecolor{myorange}{RGB}{237, 125, 49}
\definecolor{mygreen}{RGB}{112, 173, 71}
\definecolor{myblue1}{RGB}{68, 114, 196}
\definecolor{myblue2}{RGB}{91, 155, 213}
\definecolor{mygold}{RGB}{255, 192, 0}

\jmlrheading{1}{2020}{1-48}{4/00}{10/00}{meila00a}{Yingdong Hu, Liang Zhang, Wei Shan, Xiaoxiao Qin, Jin Qi, ZHenzhou Wu and Yang Yuan}

\ShortHeadings{Adversarial Data Encryption}{Hu, Zhang, Shan, Qin, Qi, Wu and Yuan}
\firstpageno{1}

\begin{document}

\title{Adversarial Data Encryption}

\author{\name Yingdong Hu \email hyd\_bupt@outlook.com \\
       \addr School of Computer Science\\
       Beijing University of Posts and Telecommunications\\
       Beijing, 10089, P.R.China
       \AND
       \name Liang Zhang \email zhang.liang.ml@biomind.ai \\
       \addr BioMind\\
       Beijing, 100070, P.R.China
       \AND
       \name Wei Shan \email weishanns@gmail.com \\
       \name Xiaoxiao Qin \email yini\_qxx@163.com \\
       \name Jing Qi \email jinqi1992@163.com \\
       \addr Department of Neurology \\
       Beijing Tiantan Hospital\\
       Capital Medical University \\
       Beijing, 100070, P.R.China
       \AND
       \name Zhenzhou Wu \email joe.wu@biomind.ai \\
       \addr BioMind\\
       Beijing, 100070, P.R.China
       \AND
       \name Yang Yuan \email yuanyang@tsinghua.edu.cn \\
       \addr Institute for Interdisciplinary Information Sciences\\
       Tsinghua University\\
       Beijing, 10089, P.R.China
       }

\editor{????}
\maketitle

\begin{abstract}
In the big data era, many organizations face the dilemma of data sharing. Regular data sharing is often necessary for human-centered discussion and communication, especially in medical scenarios. However, unprotected data sharing may also lead to  data leakage. Inspired by adversarial attack, we propose a method for data encryption, so that for human beings the encrypted data look identical to the original version,  but for machine learning methods they are misleading. 

To show the effectiveness of our method, we collaborate with the Beijing Tiantan Hospital, which has a world leading neurological center. We invite $3$ doctors to manually inspect our encryption method based on real world medical images. The results show that the encrypted images can be used for diagnosis by the doctors, but not by machine learning methods.
\end{abstract}


\input{intro.tex}

\input{prelim.tex}

\input{basic.tex}
\input{combine.tex}

\input{medical.tex}
\input{related.tex}

\input{conclusion.tex}



\acks{This work has been supported in part by the Zhongguancun Haihua Institute for Frontier Information Technology, the Institute for Guo Qiang, Tsinghua University under grant 2019GQG1002, and Beijing Academy of Artificial Intelligence.}

\bibliography{paper}

\end{document}

%% file: intro.tex
\section{Introduction}
Data sharing is a crucial and necessary component in many human-centered activities. For example, imagine a radiologist who wants to discuss Magnetic Resonance Imaging (MRI) findings with her/his research collaborators. To support her/his findings in a scientific way, she/he may need to send thousands of medical images to other experts. 

With the advancement of deep learning, data like medical images has become a valuable asset that is highly sought after. As a result, 
for many data-sensitive organizations such as hospitals, unprotected data sharing can be very risky. One of the major concerns is that
instead of just being used for intellectual discussion,
the data might be used for commercial purposes in training machine learning models without the owner's consent.

\begin{figure}[t]
\begin{center}
\includegraphics[width=11cm]{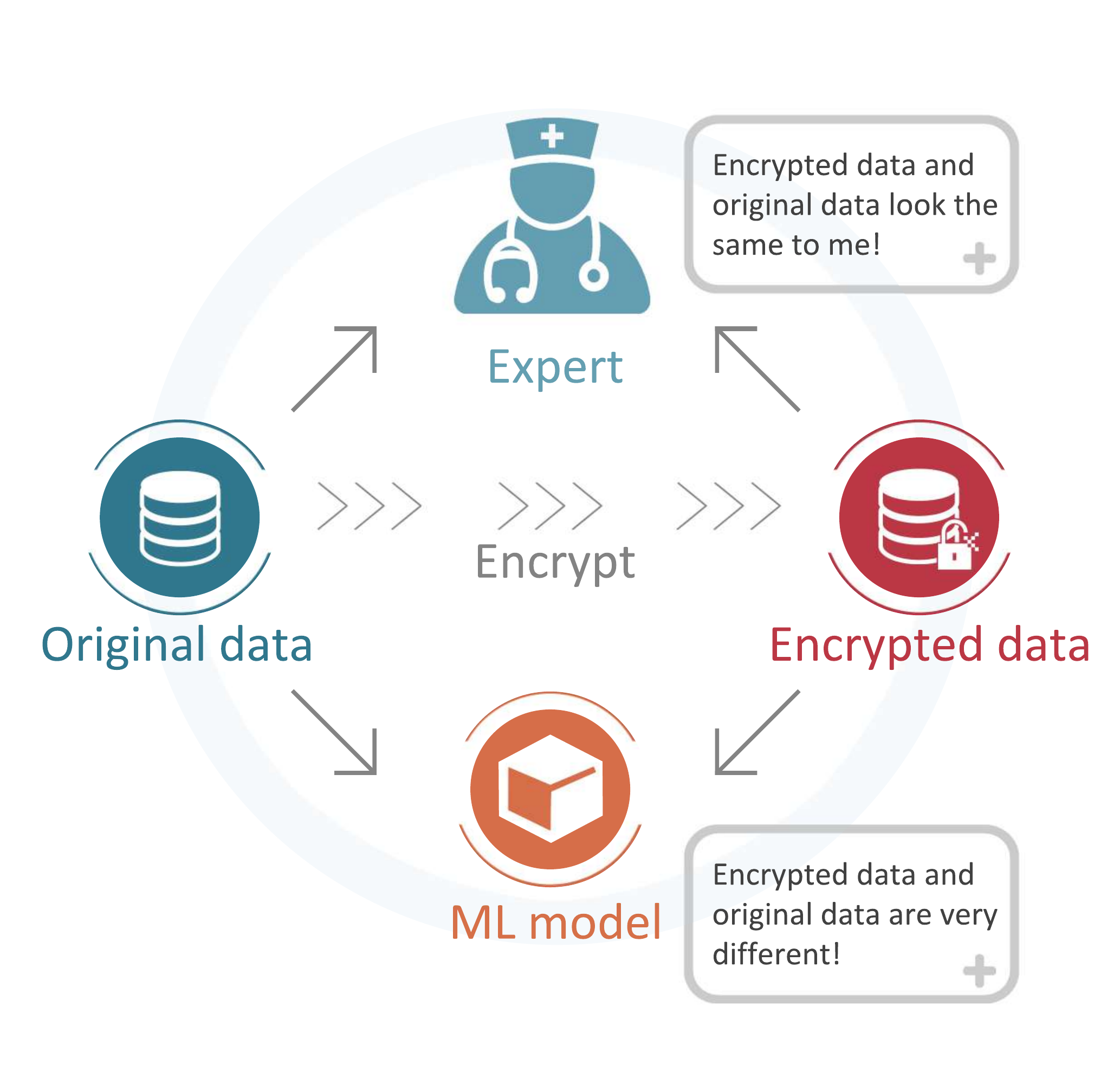}
\caption{Our goal of data sharing: for experts the encrypted data and original data can both be used for discussion, but for ML models the encrypted data are useless in training.}
\label{fig:main_goal}
\end{center}
\vskip -0.3 in
\end{figure}


Our goal is to derive a technique that encrypts the image such that it is visually indifferent from the original version, but for a model trained on the encrypted data, it works poorly on the original data as illustrated in Figure~\ref{fig:main_goal}. Doing so makes the encrypted data as well as the model trained on the encrypted data useless for the data stealers. 

We get our inspiration of encryption and content preservation from adversarial examples  \citep{Intriguing-properties-of-neural-networks,Evasion-attacks,Rotation_Translation,Synthesizing-Robust-Adversarial-Examples}, where the perturbations are visually imperceptible but can easily fool a network.  

Such adversarial examples arise because the convolutional filters tend to emphasize local features like textures or patterns \citep{brendel2019approximating}, while humans are able to focus on global structure.
Local features are non-robust because they are sensitive to minor perturbations. 
Following \citet{not-bugs}, we name the local features/global structures as non-robust/robust features, respectively.

\begin{figure*}[ht]
\begin{center}
\begin{tikzpicture}[squarednode/.style={rectangle, fill=black, very thick, minimum size=5mm}]
\begin{footnotesize}
\node at (-8,0) {\includegraphics[width=2cm]{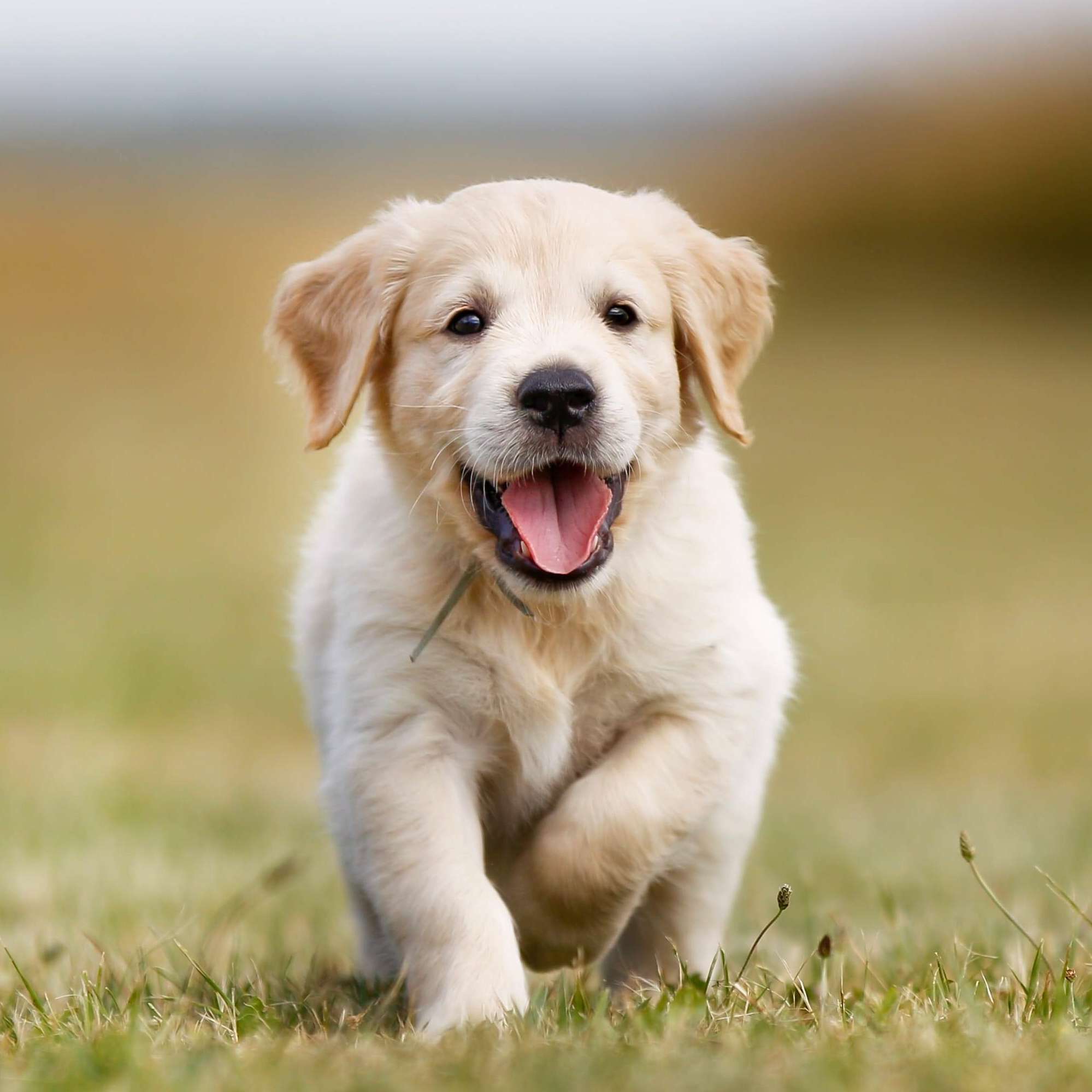}};
\node[squarednode] at (-7.25,0.80) {\color{white}dog};
\node at (-8, -1.4) {Robust features: \color{blue}Dog};
\node at (-8, -1.8) {Non-Robust features: \color{blue}Dog};
\node at (-8, 1.4) {Original training data};
\node [fill=black, single arrow, rotate=0] at (-6.1,0) {\color{white}Encrypt};
\node at (-4.0,0) {\includegraphics[width=2cm]{image/dog1.jpg}};
\node[squarednode] at (-3.25,0.80) {\color{white}dog};
\node at (-4.0, -1.4) {Robust features: \color{blue}Dog};
\node at (-4.0, -1.8) {Non-Robust features: \color{magenta}Cat};
\node at (-4.0, 1.4) {Encrypted training data};
\node [fill=black, single arrow,rotate=0] at (-2.3,0) {\color{white}Train};
\node at (-0.7,0.0) {\includegraphics[width=1.2cm]{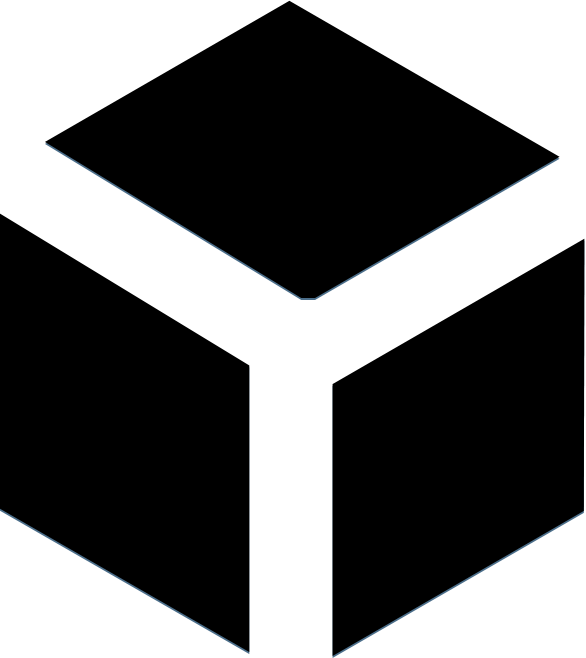}};
\node at (-0.7,1.1) {\textbf{Model}};
\node [fill=black, single arrow, rotate=35] at (0.5,1.5) {\color{white}Eval};
\node [fill=black, single arrow, rotate=-35] at (0.5,-1.5) {\color{white}Eval};
\node at (2.3,1.9) {\includegraphics[width=2cm]{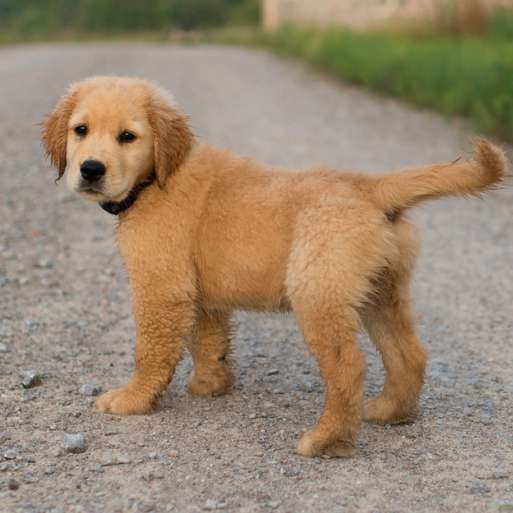}};
\node[squarednode] at (3.05,2.70) {\color{white}dog};
\node at (2.3, 0.5) {Robust features: \color{blue}Dog};
\node at (2.3, 0.1) {Non-Robust features: \color{magenta}Cat};
\node at (2.3, 3.2) {Evaluated on encrypted test set};
\node at (2.3,-1.9) {\includegraphics[width=2cm]{image/dog2.jpg}};
\node at (2.3, -3.3) {Robust features: \color{blue}Dog};
\node[squarednode] at (3.05,-1.10) {\color{white}dog};
\node at (2.3, -3.7) {Non-Robust features: \color{blue}Dog};
\node at (2.3, -0.6) {Evaluated on original test set};
\node at (4.1, 1.9) [text width=1.2cm] {\textbf{\color{green}good\\accuracy}};
\node at (4.1, -1.9) [text width=1.2cm] {\textbf{\color{red}bad\\accuracy}};
\end{footnotesize}
\end{tikzpicture}
\caption{A conceptual diagram of the encryption principle. We encrypt the data by changing the non-robust features of the data.}
\vskip -0.3 in
\label{fig:encrypt}
\end{center}
\end{figure*}

While adversarial examples were commonly used for hacking ML systems \citep{Synthesizing-Robust-Adversarial-Examples,evtimov2017robust,adversarialpatch,li2019adversarial,li2019adversarial2}, 
it turns out that we can use it as a data encryption method by exploiting the weakness of the existing models in handling local features.
See Figure~\ref{fig:encrypt} for details. 
The goal is to encrypt these local features with easy-to-learn and misleading information such that a trained network will easily overfit this misleading information, making the model prone to wrong judgments on the original images.




The security of our method relies on the hardness of solving the adversarial example problem. 
In other words, our encryption method fails if
there are models that are trained in adversarial domain but work indifferently in non-adversarial domain (or vice versa), or 
there are methods for projecting the adversarially perturbed data into the natural data manifold. However, 
there had been many explorations in 
both directions~\citep{meng2017magnet,song2017pixeldefend,samangouei2018defense,santhanam2018defending,xie2019feature}, but so far no satisfactory solution has been found.

To show the effectiveness of our method, we first conduct experiments on CIFAR-10. On top of ensuring the encryption works, we also make sure that the image before and after encryption is of negligible difference.
To further investigate whether the encryption will destroy minute local information, we collaborate with Beijing Tiantan Hospital, which has the largest  neurological center in China. 
We conduct the experiment on $5,314$ MRI images, and also invite $3$ doctors to manually 
evaluate the difference between the original and the encrypted images. 
Our results show that our encryption 
does not affect the diagnosis of the doctors, but 
significantly lower the performance of machine learning methods. 


%% file: prelim.tex
\section{Preliminaries}
We consider the standard classification setting. Let $\mathcal{X} \subseteq \mathcal{R}^{n}$ be the input domain, $\mathcal{Y}=[m]$ be the output domain, where $m$ is the number of classes. Let $(x_{i},y_{i})$ be the $i$-th instance in
the original training set, 
where  $x_i \in \mathcal{X}$ is
the input 
and $y_{i} \in  \mathcal{Y}$ is the corresponding true label. 
The encryption process is to change every pair $(x,y)$ to the encrypted pair $(x_{enc},y)$ where $x_{enc}\in \mathcal{X}$. 
In this paper, we consider the case that 
$x_{enc}$ is close to $x$ in terms of $\ell_{2}$ distance. 

The learning system aims to learn a classifier $f:\mathcal{X}\rightarrow \mathcal{Y}$ using the encrypted training set, which predicts label $y$ given input $x_{enc}$. 
We hope that the training will ``fail'': after training, $f$ has high classification accuracy on the encrypted data, but low classification accuracy on the original data. This means the training was successful on the encrypted data, but it cannot be used for natural data. 

\subsection{Robust and Non-robust Features}
In order to explain adversarial examples, people have proposed the notion of robust and non-robust features \citep{not-bugs}. 
There are no formal definitions, but intuitively, 
each data point $x$ may contain both ``robust'' and ``non-robust'' features. 
Robust features correspond to patterns that are predictive of the true label even when $x$ is adversarially perturbed under some pre-defined perturbation set, e.g. the $\ell_{2}$ ball. 
Conversely, non-robust features correspond to patterns that are also predictive, but can be easily ``flipped'' by adversarial perturbations.

Humans can only perceive robust features, so after adversarial perturbations, we can hardly see the changes in the data. 
However, ML models will use both types of features to minimize the loss during the training, therefore flipping non-robust features will have a huge impact on their prediction accuracy.

%% file: basic.tex
\section{Basic Encryption}
In this section, we introduce the basic encryption method based on adversarial attack. 
We will first present the specific steps of our encryption method and explain the underlying principle. Afterward, we run simulations on CIFAR-10 to validate our method, and also show one potential weakness of the basic encryption method. 


\subsection{Basic Encryption Method}
We first  define a permutation $P: [m]\rightarrow [m]$ for all the $m$ classes, and we call $P$ a class correspondence. For a  given class $y$, $P$ decides the target class $P(y)$ so that the inputs belong to class $y$ will be perturbed with non-robust features of class $P(y)$. 

We also need to train a base classifier $f_{\theta}$ on the original training set, where $f$ can be any modern neural network structure (e.g. ResNet or DenseNet). Given any input-label pair $(x,y)$ in the original training set, we compute the encrypted input $x_{enc}$ as follows.
\begin{equation}
    x_{enc}=\mathop{\arg\min}_{\left \|{x}'- x  \right \| \leq  \varepsilon}  \mathcal{L}(f_{\theta}({x}'), P(y))
\end{equation}

where $\mathcal{L}$ is the loss
function defined for the prediction $f_{\theta}(x')$ and target label $P(y)$, and $\varepsilon$ is a small constant. We solve this optimization problem using projected gradient descent (PGD) \citep{madry2017towards}. Because $\left \|x_{enc}- x  \right \|$ is small, the resulting encrypted inputs $x_{enc}$ looks almost the same as the original input $x$.


All the encrypted input-label pairs $(x_{enc}, y)$ make up the new encrypted training set. 
The overall encryption process is similar to the normal adversarial attack, except that \emph{we pick a fixed target class for each source class}, and the attack is made for the training set instead of the test set. As we will see in Section \ref{sec:simulation_cifar}, using a fixed target class is necessary in our setting. 

\subsection{Underlying Intuition}
\label{sec:underlying}
When applying the encryption, we change the non-robust features of the data. All the inputs in the encrypted training set exhibit non-robust features correlated with target class $P(y)$, but robust features correlated with source class $y$. 
For example, for every image containing a dog, we may modify its non-robust feature to be cat. Hence, 
during the training, 
robust dog features and non-robust cat features always come together and are labeled as ``dog''. 
Similarly, robust bird features and non-robust dog features always come together and are labeled as ``bird''. 
The classifier will easily learn this pattern. However, when a natural dog image comes, it has both robust dog features and non-robust dog features. The classifier will get confused because robust dog features are correlated with the label ``dog'', but non-robust dog features are correlated with the label ``bird''. Two guidelines confuse the model and cause it to make incorrect predictions.
A conceptual description of the encryption principle can be found in Figure~\ref{fig:encrypt}.



\subsection{Simulation of Basic Technique}
\label{sec:simulation_cifar}
We have run simulations based on CIFAR-10 dataset, which contains $10$ classes. 
Our base classifier $f_{\theta}$ is a ResNet-50, and its classification accuracy on the original test set is  $94.76\%$. We then randomly generate a class correspondence function for the source class. The correspondence $P$ we use is: $0\rightarrow8, 1\rightarrow3, 2\rightarrow1, 3\rightarrow0, 4\rightarrow 2, 5\rightarrow4, 6\rightarrow9, 7\rightarrow6, 8\rightarrow7, 9\rightarrow5$ .

During the encryption process, 
our perturbations are constrained in $\ell_{2}$-norm while each PGD step is normalized to a fixed step size. We only add small perturbation, the $\epsilon$, step size and iterations are set as 0.5, 0.1, 100 respectively. 

In Table~\ref{table:cifar-10-acc}, we 
report the performance of three different models \citep[all models are implemented in PyTorch and used in][]{bigballon2019cifarzoo}
in four different settings: Enc, Orig, Orig+P, R+Orig. 
In all settings, the models are trained using the encrypted training set, but in the first three settings, the training set is encrypted using our fixed class correspondence (fixed class correspondence is a key step in the basic encryption method), while in R+Orig, we encrypt each data point by picking random class targets. The test data set in Enc is encrypted using our class correspondence function, and for other three settings, we use the original test set. On top of that, in Orig+P, we add a post-processing step, where the output of the model is $P(f_{\theta}(x))$ for the input $x$ and classifier $f_{\theta}$.

\begin{table}[t]
\begin{center}
\begin{tabular}{l|cccc}
\hline
Model & Enc & Orig & Orig+P & R+Orig\\
\hline
DenseNet-100bc \citep{huang2019convolutional}       & 94.70\% & 22.61\% & 32.00\% & 94.26\%\\
PreResNet-110 \citep{DBLP:journals/corr/HeZR016}      & 94.64\% & 20.67\% & 41.95\% & 93.26\%\\
VGG-19 \citep{DBLP:journals/corr/SimonyanZ14a}            & 93.58\% & 28.77\% & 34.43\% & 92.17\%\\
\hline
\end{tabular}
\caption{Classification accuracies of different models on the CIFAR-10 encrypted test set and original test set.}
\label{table:cifar-10-acc}
\end{center}
\end{table}

\begin{table}[t]
\vskip -0.1in
\begin{center}
\begin{tabular}{cc}
\hline
Class & Proportion \\
\hline
0           &  9.4\% \\
1           &  0.3\% \\
2           &  0.2\% \\
3           &  36.8\% \\
4           &  30.6\% \\
5           &  0.9\% \\
6           &  0.3\% \\
7           &  0.2\% \\
8           &  21.2\% \\
9           &  0.1\% \\
\hline
\end{tabular}
\caption{The prediction distribution for images in class $0$.}
\label{table:detail-for-0}
\end{center}
\vskip -0.1in
\end{table}

From the first two columns of the Table~\ref{table:cifar-10-acc} (Enc and Orig), we can see that our encrypted dataset has achieved its purpose:
the model has high accuracy on the encrypted test set (the accuracy is similar to the results obtained by training on the original training set and predicting on the original test set), and has extremely low accuracy on the original test set.
We remark that the high accuracy on the encrypted test set does not mean that the model will be useful if one knows how to encrypt the data. This is because when creating an encrypted data set, we need the extra information of the correct label of each input, and then apply the correspondence function $P$ to perturb the input. In practice, without knowing the true label, one cannot encrypt the data even with the corresponding function $P$. In other words, the accuracies for Enc is vacuous in practice, and we include it here to serve as a strong benchmark for comparison. 

If the model relies more on robust features, the accuracies in Orig (second column in Table~\ref{table:cifar-10-acc}) should not be extremely low. If the model relies more on non-robust features, the accuracies in Orig+P (third column in Table~\ref{table:cifar-10-acc}) should be much higher than Orig. However, we see accuracies in Orig are very low, and accuracies in Orig+P are slightly higher than Orig. This shows 
that the trained model  gets confused when seeing the original images (as we explained in Section~\ref{sec:underlying}), so it may make predictions different from $y$ or $P(y)$. Indeed, as we show in Table~\ref{table:detail-for-0}, the prediction distribution of the trained model for images in class $0$ is not well concentrated in class $0$ or class~$8$ (the target class of 0 is 8).

Moreover, the accuracies in R+Orig show that using fixed target class is very necessary, otherwise the trained model will have equally good accuracy as the normal case, which means the encryption fails. 

\subsection{Recovery of Correspondence Function $P$}
\label{sec:recovery-p}
As we mentioned before, decrypting the encrypted data is as hard as solving the adversarial example problem. 
Instead of directly decrypting the data and cracking our method, in this subsection we consider a simpler problem of recovering the correspondence function $P$. 
According to Table~\ref{table:cifar-10-acc},
with the recovery of $P$ the attacker can only slightly increase the accuracy, but in practice that is also the extra hidden information that we do not want to share. 

Assume that there exists an attacker who knows our encryption method, and also has the encrypted data set $D$ (e.g., the CIFAR-10 dataset we used in the previous subsection). He may also have some different labeled data $D'$ sampled from the original data distribution, obtained from other sources. 

He can learn the secret class correspondence function as follows.
First, he  trains a 
classifier $M_1$ on $D$ and a classifier $M_2$ on $D'$. 
Then he uses $M_2$ to simulate our encryption process, that is, based on a correspondence function $P$, he  modifies the 
the data points in $D'$ to have incorrect non-robust features according to $P$. After encrypting $D'$, he uses $M_1$ to make predictions on it. Usually, we assume $P$ is a permutation, but here we relax this requirement and allow the target class to appear in multiple locations. Now the attacker can fix the target class for classes $1$-$9$, and enumerate the target class for class $0$.  For each possible correspondence $P$, he evaluates the performance of $M_1$, and finally picks the one with the highest accuracy for class $0$, which will be the actual target class in $P$ (see Table~\ref{table:basic-attack}). After processing the images of classes $1$-$9$ in this way, the attacker can get the correct correspondence within $100$ attempts.

\begin{table}[t]
\vskip -0.1in
\begin{center}
\begin{tabular}{lc}
\hline
Test set & Test set acc \\
\hline
$0\rightarrow 0$ (orig test set)       &  22.61\% \\
$0\rightarrow 1$                       &  22.15\% \\
$0\rightarrow 2$                       &  22.10\% \\
$0\rightarrow 3$                       &  22.11\% \\
$0\rightarrow 4$                       &  22.21\% \\
$0\rightarrow 5$                       &  22.17\% \\
$0\rightarrow 6$                       &  22.24\% \\
$0\rightarrow 7$                       &  22.22\% \\
$0\rightarrow 8$                       &  \textbf{31.49\%}\\
$0\rightarrow 9$                       &  22.30\% \\
\hline
\end{tabular}
\caption{We simulate the attack process. Suppose we use the correspondence $P$ defined in Section \ref{sec:simulation_cifar} (i.e. $0\rightarrow 8$). Assume the attack has a PreResNet-20 trained on the CIFAR-10 encrypted training set as classifier $M_1$, and has a ResNet-50 trained on CIFAR-10 original training set as classifier $M_2$. The attacker fixes the target class for classes $1$-$9$, and enumerates the target class for class $0$. 
This table shows the 10 accuracies he gets from $M_1$, where $0\rightarrow 8$ has the highest accuracy.}
\label{table:basic-attack}
\end{center}
\vskip -0.3in
\end{table}


%% file: combine.tex
\section{Combined Encryption}
For the basic encryption method, the correspondence function is easy to recover, because if each class of data only corresponds to one specific class,  the number of correspondences is limited.
In this section, we present the combined encryption method, which addresses this problem. Therefore, it is not only hard to decrypt the data, but also hard to recover the specific encryption method. 

\begin{figure*}[t]
\begin{center}
\begin{tikzpicture}[squarednode/.style={rectangle, fill=black, very thick, minimum size=5mm}, plusnode/.style={circle, draw=black!70, fill=black!10, very thick, minimum size=5mm}]
\begin{footnotesize}
\node at (-8,0) {\includegraphics[width=2cm]{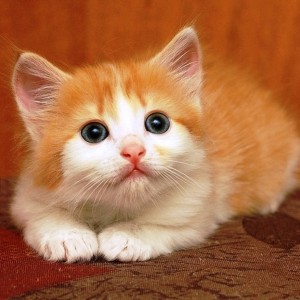}};
\node [fill=black, single arrow, rotate=30] at (-6.2,1.2) {aaaa};
\node [fill=black, single arrow, rotate=-30] at (-6.2,-1.2) {aaaa};
\node at (-2.8,1.9) {\includegraphics[width=2cm]{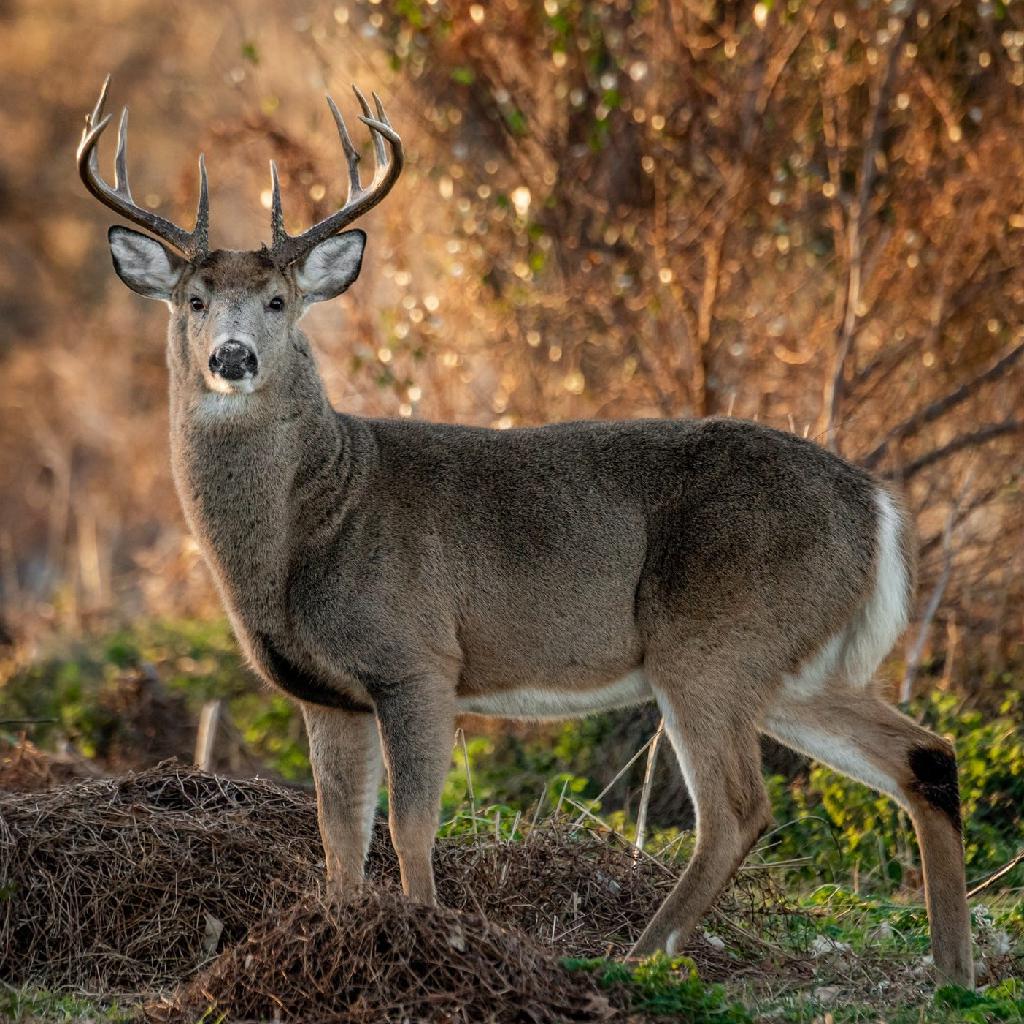}};
\node at (1.4,1.9) {\includegraphics[width=2cm]{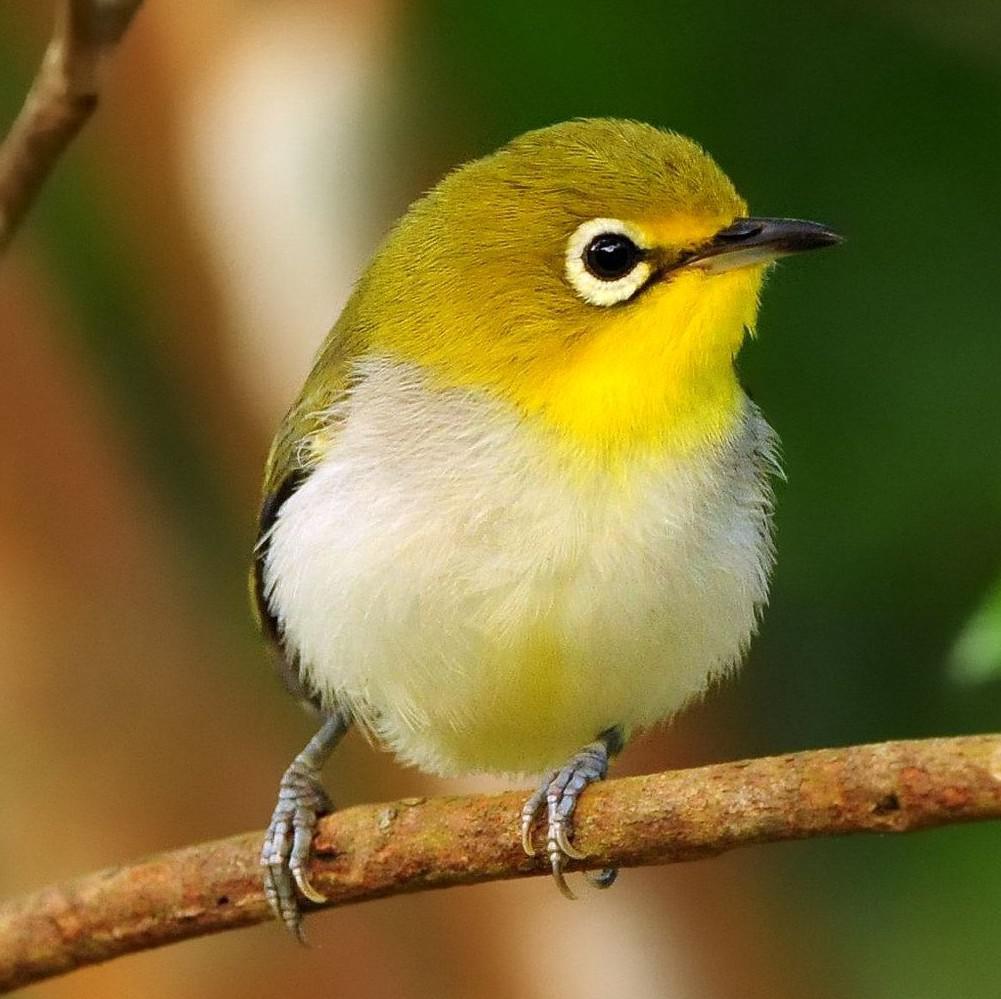}};
\node[opacity=0.5] at (-2.8,-1.9) {\includegraphics[width=2cm]{image/deer.jpg}};
\node[opacity=0.5] at (1.4,-1.9) {\includegraphics[width=2cm]{image/bird.jpg}};
\node at (5,1.9) {\includegraphics[width=2cm]{image/cat.jpg}};
\node at (5,-1.9) {\includegraphics[width=2cm]{image/cat.jpg}};

\node at (-8, 1.6) [text width=2cm,align=center]{Original training data};
\node at (-2.8, 3.5) [text width=3.0cm,align=center]{Left half non-robust features of deer};
\node at (1.4, 3.5) [text width=3.2cm,align=center]{Right half non-robust features of bird};
\node at (-2.8, -0.3) [text width=3.4cm,align=center]{$50\%$ of total non-robust features of deer};
\node at (1.4, -0.3) [text width=3.4cm,align=center]{$50\%$ of total non-robust features of bird};
\node at (5, 3.5) [text width=3.4cm,align=center]{Horiz. Concat. encrypted data};
\node at (5, -0.3) [text width=3.4cm,align=center]{Mixup \\encrypted data};

\draw[pattern=horizontal lines light gray] (-2.79,2.89) rectangle (-1.81,0.91);
\draw[pattern=horizontal lines light gray] (0.41,2.89) rectangle (1.39,0.91);
\node[squarednode] at (-7.25,0.85) {\color{white}cat};
\node[squarednode] at (-2.05,2.75) {\color{white}deer};
\node[squarednode] at (2.15,2.75) {\color{white}bird};
\node[squarednode] at (-2.05,-1.05) {\color{white}deer};
\node[squarednode] at (2.15,-1.05) {\color{white}bird};
\node[squarednode] at (5.75,2.75) {\color{white}cat};
\node[squarednode] at (5.75,-1.05) {\color{white}cat};
\node[plusnode] at (-4.9,1.9) {\begin{LARGE}\textbf{+}\end{LARGE}};
\node[plusnode] at (-4.9,-1.9) {\begin{LARGE}\textbf{+}\end{LARGE}};
\node[plusnode] at (-0.7,1.9) {\begin{LARGE}\textbf{+}\end{LARGE}};
\node[plusnode] at (-0.7,-1.9) {\begin{LARGE}\textbf{+}\end{LARGE}};
\node [fill=black, single arrow, rotate=0] at (3.2,1.9) {aaaa};
\node [fill=black, single arrow, rotate=0] at (3.2,-1.9) {aaaa};
\end{footnotesize}
\end{tikzpicture}
\caption{Combined encryption method with different filters. The first row shows the Horizontal Concat method: the left 50\% of image 1 is horizontally concatenated with the 50\% of image 2. The second row shows the Mixup method: creates a new encrypted image by interpolating between two images, using a constant alpha 0.50.}
\label{fig:combined_technique}
\end{center}
\vskip -0.2in
\end{figure*}

\subsection{Combined Encryption Method}
For the combined encryption methods, each class $y$ corresponds to multiple target classes (i.e. we select multiple target classes $\left \{ t^y_{i} \right \}_{i=1}^{N(y)},N(y)\geq2$). Then we modify each input-label pair $(x,y)$ from the original training set as follows. For each target class $t^y_{i}$, construct encrypted input $x_{i}$ using the basic encryption method described above. Then, we combine all the encrypted input $\left \{ x_{i} \right \}_{i=1}^{N(y)}$ into a new encrypted input $\tilde{x}$. Formally:
\begin{equation}
    \tilde{x}=T\left(\{x_{i}\}_{i=1}^{N(y)}\right )\quad N(y)\geq2
\end{equation}
where $T$ is a function that maps multiple inputs into a single new encrypted input, 
and at the same time keeps $\left \|\tilde{x}- x  \right \|$ small. In other words, humans can hardly notice the difference between $\tilde{x}$ and $x$.
For example, $T$ may be a function that concatenates two examples horizontally, or generates a mixing coefficient and produces the new example as a convex combination \citep[similar to Mixup][]{mixup}. 
See Figure~\ref{fig:combined_technique} for illustration. 

The search space of our combined encryption method is exponentially larger than the basic encryption method. 
First, each class of data corresponds to multiple classes (the number $N(y)$ can vary for different classes), which greatly increases the number of class correspondences.
Secondly, 
there are many different valid $T$ functions for combining data. For example, various data augmentation methods such as Mixup~\citep{mixup}, CutMix~\citep{cutmix}, and Random square~\citep{8659168} can be adapted as methods to combine multiple inputs into new encrypted inputs. As a result, it is very difficult for an attacker to recover the encryption method (see detailed explanation in Section \ref{sec:hard-decrypt}).


\subsection{Simulation Results}
In this subsection, we demonstrate two combined encryption methods on CIFAR-10 dataset: 1) Horizontal Concat; 2) Mixup And Concat.

\subsubsection{Horizontal Concat}

We select two target classes for each source class (as shown in Table~\ref{table:encrypt-corres}), then use  PGD to add adversarial perturbations to the image based on its two targets. For each image, we get two sightly changed images. The left $50\%$ of Image 1 is horizontally concatenated with the $50\%$ of Image 2. In practice, we may pick other composition ratios for each source class, e.g., $80\%$ and $20\%$.
As a result, each image contains non-robust features of two target classes.

\begin{table}[t]
\begin{center}
\begin{tabular}{lcccccccccc}
\hline
class      & 0 & 1 & 2 & 3 & 4 & 5 & 6 & 7 & 8 & 9  \\
target1    & 8 & 3 & 1 & 0 & 2 & 4 & 9 & 6 & 7 & 5  \\
target2    & 4 & 2 & 3 & 5 & 7 & 1 & 8 & 0 & 6 & 9  \\
\hline
\end{tabular}
\caption{Encrypt correspondence used in Horizontal Concat.}
\label{table:encrypt-corres}
\end{center}
\end{table}

\begin{table}[t]
\begin{center}
\begin{tabular}{lc}
\hline
Test set & Test set acc \\
\hline
orig test set                      &  31.70\% \\
50\%left + 50\%right(encrypted)    &  94.44\% \\
40\%left + 60\%right               &  94.10\% \\
30\%left + 70\%right               &  92.26\% \\
20\%left + 80\%right               &  90.28\% \\
10\%left + 90\%right               &  85.96\% \\
100\%right                         &  82.79\% \\
\hline
\end{tabular}
\caption{Classification accuracies obtain by changing the composition ratio of two target images}
\label{table:changing_portion}
\end{center}
\vskip -0.3in
\end{table}
One benefit of this method is that the ratio between the two images is unknown to the attacker (not necessarily $50\%$-$50\%$), so it
provides protection for the encryption process. 
As described in Section \ref{sec:recovery-p}, if the attacker wants to recover the correspondence function, 
he needs to know not only which classes each type of image corresponds to, but also how the two pictures are concatenated together.  Table~\ref{table:changing_portion} shows the accuracy obtained by using the correct target class set but different composition ratio, set fixed for all source classes (this is a simplification, because empirically one may need to pick different composition ratio for different source classes). Hence, it is no longer easy for the attacker to pick the correct correspondence class set by only looking at the relationship between accuracy and correspondence class, because the composition ratio has a great impact on the final accuracy.


\begin{table}[t]
\begin{center}
\begin{tabular}{lcccccccccc}
\hline
class      & 0 & 1 & 2 & 3 & 4 & 5 & 6 & 7 & 8 & 9  \\
target1    & 8 & 3 & 1 & 0 & 2 & 4 & 9 & 6 & 7 & 5  \\
target2    & 4 & 2 & 3 & 5 & 7 & 1 & 8 & 0 & 6 & 9  \\
target2    & 6 & 2 & 9 & 1 & 0 & 7 & 5 & 3 & 4 & 8  \\
target2    & 3 & 9 & 6 & 1 & 5 & 8 & 4 & 7 & 0 & 2  \\
\hline
\end{tabular}
\caption{Encrypt correspondence used in Horizontal Mixup And Concat.}
\label{table:encrypt_4_targets}
\end{center}
\end{table}

\begin{table}[t]
\begin{center}
\begin{tabular}{lcc | cc}
\hline
Model & Enc & Orig & Enc & Orig\\
\hline
DenseNet-100bc    & 94.62\% & 29.69\% & 94.45\% & 32.92\%\\
PreResNet-110     & 94.49\% & 32.65\% & 94.03\% & 37.21\%\\
VGG-19            & 94.30\% & 48.13\% & 93.06\% & 55.00\%\\
\hline
\end{tabular}
\caption{Classification accuracies of different models on the CIFAR-10 encrypted test set and original test set when using Horizontal Concat and Mixup And Concat. The first and second columns correspond to Horizontal Concat, the third and fourth columns correspond to Mixup And Concat.}
\label{table:combine-technique-acc}
\end{center}
\vskip -0.3in
\end{table}

\begin{table}[t]
\begin{center}
\begin{tabular}{lc}
\hline
Test set & Test set acc \\
\hline
orig test set                        &  33.29\% \\
(8,4)                                &  41.31\% \\
(4,8)                                &  41.28\% \\
(8,6)                                &  40.37\% \\
(8,5)                                &  40.76\% \\
(8,3)                                &  39.82\% \\
(1,4)                                &  33.06\% \\
(2,4)                                &  38.53\% \\
(3,4)                                &  36.50\% \\
(9,8)                                &  34.81\% \\
(7,8)                                &  39.83\% \\
(4,2)                                &  38.16\% \\
(4,3)                                &  36.45\% \\
(7,0)                                &  32.85\% \\
(5,0)                                &  32.90\% \\
(9,2)                                &  33.51\% \\
(0,3)                                &  32.60\% \\
(6,0)                                &  32.52\% \\
\hline
\end{tabular}
\caption{Classification accuracies by adding two non-robust features to class 0 image}
\label{table:samples}
\end{center}
\end{table}

\begin{table}[t]
\vskip -0.1in
\begin{center}
\begin{tabular}{lc}
\hline
Test set & Test set acc \\
\hline
mixup and concat(encrypted)          &  93.85\% \\
orig test set                        &  48.14\% \\
target1                              &  81.53\% \\
target1+target2(horiz concat)        &  82.50\% \\
\hline
\end{tabular}
\end{center}
\vskip -0.1in
\caption{Classification accuracy by guessing different number of targets}
\label{table:guessing_targets}
\end{table}

\subsubsection{Mixup And Concat}
Mixup And Concat is a more complicated combination method, where each class of image corresponds to four classes, as shown in the Table~\ref{table:encrypt_4_targets}.  Similar to the previous method, for each image we first get four new target images. Then we mixup Image 1 and Image 2 to get Image 5 (e.g. $\mathrm{Image~}5 = 0.4 * \mathrm{Image~} 1 + 0.6 * \mathrm{Image~} 2$), mixup Image 3 and Image 4 to get Image 6 (e.g. $\mathrm{Image~} 6 = 0.6 * \mathrm{Image~} 3 + 0.4 * \mathrm{Image~} 4$). Finally, the left $50\%$ of Image 5 is horizontally concatenated with the right $50\%$ of Image 6.

Compared with Horizontal Concat, 
this approach is more secure. If the attacker does not know the encryption correspondence in advance, it is difficult for him to figure out the encryption method, as we will see in the next section.  Table~\ref{table:combine-technique-acc} shows that both Horizontal Concat and Mixup And Concat methods work well, in the sense that after encryption, the data can no longer be used for ML training. 


\subsection{Recovery of the Encryption Method}
\label{sec:hard-decrypt}

To show it is difficult to recover the encryption method, 
we start with Horizontal Concat, 
where each source class corresponds to two other classes, and the encrypt correspondence is defined in Table \ref{table:encrypt-corres}. 
As in Section \ref{sec:recovery-p},
we assume that the attacker has the encrypted dataset and some labeled original data. To find out the encryption method, he first needs to decide $N(y)$ for each source class $y$, which is very difficult. 


Assume he can set $N(y)=2$ for all $y$, as we used in our encryption, and he needs to figure out the specific correspondence function. For example, as in Section \ref{sec:recovery-p}, we may start with source class $0$.
It may correspond to one hundred combinations ( e.g. $0\rightarrow(6,3) , 0\rightarrow(4,5)$, etc). Table~\ref{table:samples} shows some combinations and their accuracy, assuming the encryption uses $50\%$ and $50\%$ composition ratio. After trying all $100$ combinations, the attacker will find the two with the highest accuracy ($0\rightarrow(4,8) , 0\rightarrow(8,4)$), which is the correct correspondence. However, 
from classes $1$-$9$, he has to repeat this process and tries a total of $1,000$ times to find all the target classes. This overhead is quite large, and this is just the case when Horizontal Concat is used. If each class image corresponds to more classes, such as using Mixup and Concat, he cannot find the correspondence efficiently. 
See Table~\ref{table:guessing_targets}, where if Mixup and Concat is used, even the attacker knows the exact composition ratio and correspondence function, he can only get $82.50\%$ test accuracy compared with $93.85\%$ if he knows the exact encryption method. Moreover, $82.50\%$ is fairly close to the accuracy of $81.53\%$, which is the accuracy one can get if he knows one of the target classes used in Mixup and Concat. 

In addition, the various hyperparameters in the encryption process (the value of $\epsilon$, step size, iteration), how multiple images are combined (Mixup, Concat, or CutMix) are also unknown to the attacker. All these factors show that it is hard for the attacker to even recover the encryption method.

\subsection{Domain Adaptation}
Although the encrypted data cannot be used for training, maybe it can still provide some other extra information about the data distribution because the encrypted data look similar to the original data. However, in this subsection, we demonstrate that the encrypted data are from a distribution different from the natural data distribution, by applying techniques from domain adaptation. 

The rationale is the following. If the two datasets are close to each other, then training on the first one and testing on the second one will give us good accuracy. However, if the two datasets are far away from each other, the training on the first one and testing on the second one will give us bad accuracy. In the latter case, we may use domain adaptation methods to improve the test accuracy. 

In Table \ref{table:GTA}, we use Generate To Adapt (GTA) described in \cite{Sankaranarayanan_2018_CVPR} to illustrate the idea. 
To form a comparative experiment, we test the effect of using GTA on original training set and new CIFAR-10 test set (as described in \citet{DBLP:journals/corr/abs-1806-00451}, there exists a small distribution shift between the original CIFAR-10 dataset and the new test set). From Table~\ref{table:GTA}, we observe that GTA method improves the accuracy of new test set from $83.80\%$ to $89.28\%$. This proves that GTA can solve the general domain shift problems. But it can only help the attacker improve the accuracy of the original test set from $28.73\%$ to $49.72\%$. This shows that GTA has little effect on our encrypted dataset. In other words, the encrypted data distribution and the original data distribution are far away from each other. Attempts to crack our encryption methods are much more complex and difficult to solve than the general domain shift problem.

\begin{table}[t]
\vskip 0.15in
\begin{center}
\centering
\begin{tabular}{l|cc}
\hline
Method &  orig train set$\rightarrow$new test set  & enc train set$\rightarrow$orig test set \\
\hline
source only   &83.80\%   &28.73\%   \\
GTA           &89.28\%   &49.72\%   \\    
\hline
\end{tabular}
\caption{Classification accuracies of GTA on original CIFAR-10 and encrypted CIFAR-10}
\label{table:GTA}
\end{center}
\vskip -0.3in
\end{table}

%% file: medical.tex
\section{Real World Experiments on Medical Data}

\begin{figure*}[t]
\begin{center}
\includegraphics[width=\textwidth]{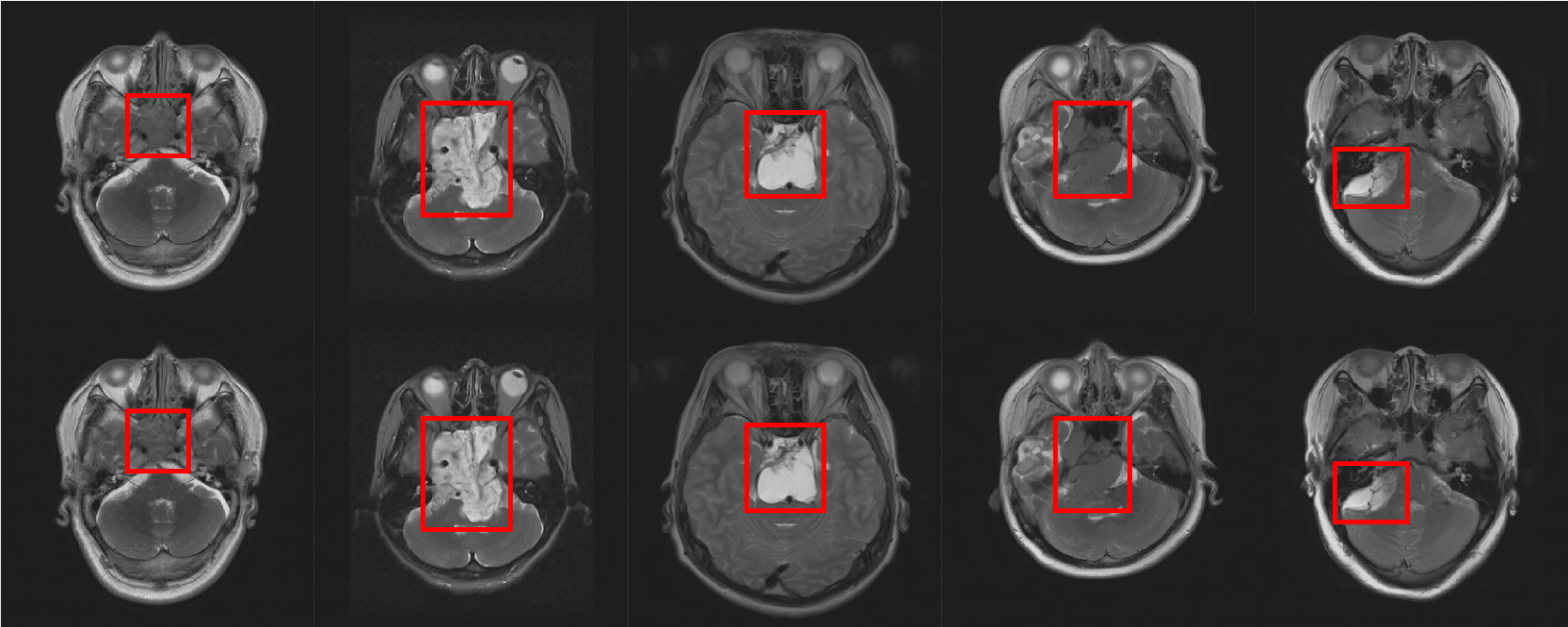}
\caption{Brain MRIs from the original test set (top row) and corresponding images from the encrypted test set (bottom row). For each original image, the tumor is marked out using red bounding box.}
\label{fig:brain-image}
\end{center}
\vskip -0.3in
\end{figure*}

Magnetic Resonance Imaging (MRI) provides excellent soft tissue contrast of brain tumors without exposing the patient to radiations, consequently it is widely used in the clinical diagnosis of brain tumors. We collaborate with a world-leading neurological center and use their preprocessed, isotropic interpolated brain MRIs as experiment data. For each MRI, we choose the cross-section with the largest tumor size and form a 2D image dataset. The dataset consists of $5,314$ MRIs (320$\times$320 pixels each) in 5 classes, and the number of images in each class is not the same. 20\% of images from each class is selected as the test set, and the remaining images are used as the training set (As shown in Table~\ref{brain_tumor_dataset}).

\begin{table}[t]
\vskip 0.15in
\begin{center}
\begin{tabular}{l|cc}
\hline
Class & Number in training set & Number in test set\\
\hline
Meningioma         & 879  & 219 \\
Chordoma            & 1149 & 287 \\
Schwannoma          & 541  & 135 \\
Pituitary Adenoma   & 1451 & 362 \\
Craniopharyngioma   & 233  & 58  \\
\hline
Total               & 4253 & 1061 \\
\hline
\end{tabular}
\caption{Number of each class of brain tumor MRI in the dataset.}
\label{brain_tumor_dataset}
\end{center}
\vskip -0.1in
\end{table}

\begin{table}[ht]
\vskip 0.05in
\begin{center}
\begin{tabular}{lcccr}
\hline
Architecture & Enc test set & Orig test set \\
\hline
DenseNet-100bc & 79.01\% & 38.84\% \\
PreResNet-110  & 79.92\% & 33.14\% \\
PreResNet-20   & 78.10\% & 43.64\% \\
VGG-19        & 80.01\% & 48.84\% \\
VGG-11        & 79.98\% & 50.15\% \\
\hline
\end{tabular}
\caption{Classification accuracies of different models on the brain tumor MRI encrypted test set and original test set.}
\label{table:brain-tumor-acc}
\end{center}
\vskip -0.3in
\end{table}

\tikzset{
module/.style={rounded corners, align=center, thick},
module1/.style={module, top color=red!10, bottom color=red!35, draw=red!75, text width=25mm, minimum height=15mm},
module2/.style={module, top color=cyan!10, bottom color=cyan!35, draw=cyan!75, text width=25mm, minimum height=15mm},
module3/.style={module, top color=green!10, bottom color=green!35, draw=green!75, text width=25mm, minimum height=15mm},
module4/.style={module, top color=yellow!10, bottom color=yellow!35, draw=yellow!75, text width=25mm, minimum height=15mm},
}
\begin{figure*}[ht]
\vskip 0.2 in
\begin{center}
\begin{tikzpicture}
\node at (-7,1.3) [module1] {Original MRIs};
\node at (-7,-1.3) [module1] {Encrypted MRIs};
\node at (-2.5,1.3) [module2] {Diagnosis};
\node at (-2.5,-1.3) [module2] {Diagnosis};
\node at (1.4,0) [module3] {Diagnosis distribution};
\node at (5,0) [module4] {Error assay};
\draw[->, red!60, ultra thick] (-7, 0.5) --(-7,-0.5);
\draw[->, red!60, ultra thick] (-5.5, 1.3) --node [pos=.5, above, sloped] {Shuffle} (-4.0,1.3);
\draw[->, red!60, ultra thick] (-5.5, -1.3) --node [pos=.5, above, sloped] {Shuffle} (-4.0,-1.3);
\draw[->, cyan!60, ultra thick] (-1.0, 1.3) -- (-0.1,0.1);
\draw[->, cyan!60, ultra thick] (-1.0, -1.3) -- (-0.1,-0.1);
\draw[->, green!60, ultra thick] (2.9, 0) -- (3.5, 0);
\node at (-6.3,0) {\color{red!60}Encrypt};
\end{tikzpicture}
\vskip 0.1 in
\caption{Doctor's evaluation process for original MRIs and encrypted MRIs}
\label{fig:diagnosis-process}
\end{center}
\vskip -0.1 in
\end{figure*}
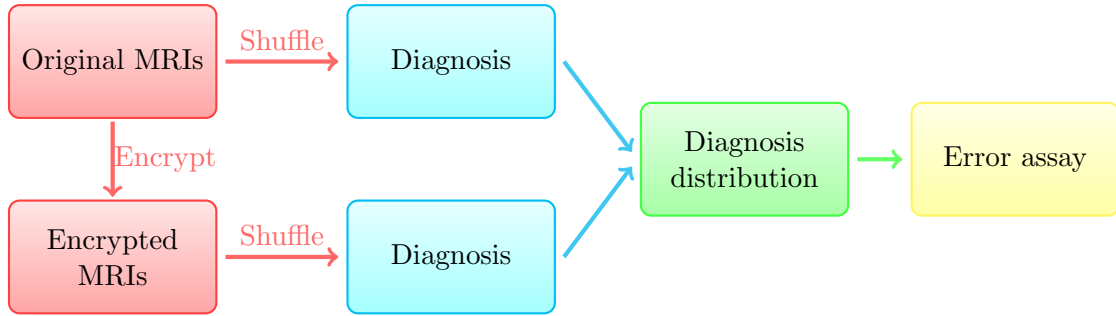

\begin{figure}[ht]
\vskip 0.2 in
\begin{center}
\begin{tikzpicture}
\pie [pos ={-2,0}, radius=1.8, rotate=10,text = inside, color = {myorange, myblue1, mygreen}] {72/Correct, 23/Incorrect, 5/Errors};
\pie [pos ={2.5,0}, radius=0.8, rotate=-10, sum=5, color = {myblue2, mygold}] {3/, 2/};
\node at (2.5, 1.4) [text width=4cm,align=center]{\small Correct on encrypted\\Wrong on original};
\node at (2.5, -1.4) [text width=4cm,align=center]{\small Correct on original\\Wrong on encrypted};
\draw[color=black!50] (-0.2,0.31) -- (2.2,0.75);
\draw[color=black!50] (-0.2,-0.29) -- (2.2,-0.75);
\node at (2.682, 0.24) {\%};
\node at (2.76, -0.33) {\%};
\end{tikzpicture}
\caption{Results of doctors' diagnoses on the original test set and the encrypted test set.}
\label{fig:diagnosis-result}
\end{center}
\vskip -0.1in
\end{figure}
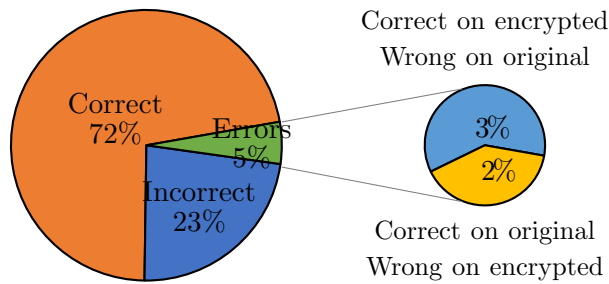

We use the Horizontal Concat encryption technique. The base classifier is ResNet-18 (since the dataset is not huge, we choose a smaller model to avoid overfitting), and its classification accuracy on the original test set is 79.92\%. The correspondence $P$ we used is: $0\rightarrow(1,3), 1\rightarrow(2,4), 2\rightarrow(3,0), 3\rightarrow(4,1), 4\rightarrow(0,2)$. The adversarial perturbations are constrained in $\ell_{2}$-norm. Hyperparameter $\epsilon$, step size and iterations are set as 2, 0.1, 100 respectively.  The experimental process is similar to that in Section 3.3. After constructing the encrypted dataset using the Horizontal Concat encryption encryption method, we train different models on the encrypted training set and observe their performances on the encrypted test set and the original test set. 

Results are shown in Table~\ref{table:brain-tumor-acc}. 
We can see that all the models trained on the encrypted training set have extremely low accuracies on the original test set. We also sample some images from the original and encrypted brain image test sets and display them in Figure~\ref{fig:brain-image}. 

We invite 3 doctors from Beijing Tiantan Hospital, who are experts on brain MRIs, to evaluate the difference between the original and encrypted brain image test sets. The entire evaluation process can be seen Figure~\ref{fig:diagnosis-process}. Firstly, images in the original test set and the encrypted test set are shuffled randomly. Then, doctors examine and make a diagnosis based on each image in the original test set. Thereafter, doctors examine and make a diagnosis based on each image in the encrypted test set. 

In the end, diagnoses of each patient from the two test sets are compared, and the results are summarised in Figure~\ref{fig:diagnosis-result}. 
In $95\%$ cases, doctors make the same diagnoses for both original and encrypted images, with $72\%$ cases being correct diagnoses and $23\%$ being wrong diagnoses (this is similar to other brain MRI . In the remaining $5\%$ cases, doctors make different diagnoses, with $2\%$ being correct on the original image and wrong on the encrypted image, and $3\%$ being wrong on the original image and correct on the encrypted image. These results indicate that encryption does not affect the doctors' diagnosis of brain tumors.

Overall, we think our encryption method works well on real medical data and achieves its goal: for humans the encrypted data and original data can both be used for diagnosis, but for machine learning models the encrypted data are useless in training. 

%% file: related.tex
\section{Related Work}
Researchers have been studying the data sharing problem for a long time. For example, multi-party computation~\citep{yao1982protocols,yao1986generate,goldreich2019play,chaum1988multiparty,ben2019completeness,bogetoft2009secure} considers the setting that multiple parties jointly compute a function, without the need of revealing each other's private inputs.  As another example, differential privacy~\citep{dwork2004privacy,blum2005practical, dwork2006calibrating,dwork2008differential,abadi2016deep} considers the mechanism design problem for database privacy, where adding or removing any single element in the database will only slightly change the outcome for the query to the database. 

Adversarial example is an active research area in deep learning. There have been many attack strategies to fool the neural networks~\citep{Intriguing-properties-of-neural-networks,43405,madry2017towards,Dong_2018_CVPR,Moosavi-Dezfooli_2016_CVPR,7958570}. On the other side,  researchers have tried to propose defense mechanisms against such attacks to train robust networks \citep{gu2014towards,madry2017towards,Zheng_2016_CVPR,samangouei2018defensegan,DBLP:journals/corr/abs-1805-09190,cohen2019certified,lee2019stratified}.
There are also many papers 
proposing models to explain adversarial examples~\citep{not-bugs, DBLP:journals/corr/abs-1801-02774,DBLP:journals/corr/abs-1802-08686,ford2019adversarial, DBLP:journals/corr/TanayG16,shafahi2018are,DBLP:journals/corr/abs-1809-03063,bubeck2018adversarial}, among which \citet{not-bugs} propose that adversarial perturbations arise as well-generalizing, yet brittle, features (non-robust features). 



%% file: conclusion.tex
\section{Conclusion}
In this paper, we present a new encryption method for the data-sharing problem, so that the encrypted data can be used for human-centered activities, but not for machine learning training purposes. Using the encrypted data, the data stealers cannot train a model that generalizes to original natural data. Our method is based on adversarial attack and can be divided into basic encryption method and combined encryption method. The basic encryption method solves our data sharing problem, and the combined encryption method further improves its security. We present a series of simulations on CIFAR-10 to validate both methods. We also apply our combined encryption to the real-world clinical data and find that our encryption does not affect the doctors' diagnosis of brain tumors. Our method heavily relies on the hardness of adversarial examples. Hence, for future work, it would be interesting to understand the limitation of adversarial examples theoretically.

%% file: main.bbl
\begin{thebibliography}{53}
\providecommand{\natexlab}[1]{#1}
\providecommand{\url}[1]{\texttt{#1}}
\expandafter\ifx\csname urlstyle\endcsname\relax
  \providecommand{\doi}[1]{doi: #1}\else
  \providecommand{\doi}{doi: \begingroup \urlstyle{rm}\Url}\fi

\bibitem[Abadi et~al.(2016)Abadi, Chu, Goodfellow, McMahan, Mironov, Talwar,
  and Zhang]{abadi2016deep}
Martin Abadi, Andy Chu, Ian Goodfellow, H~Brendan McMahan, Ilya Mironov, Kunal
  Talwar, and Li~Zhang.
\newblock Deep learning with differential privacy.
\newblock In \emph{Proceedings of the 2016 ACM SIGSAC Conference on Computer
  and Communications Security}, pages 308--318, 2016.

\bibitem[Athalye et~al.(2018)Athalye, Engstrom, Ilyas, and
  Kwok]{Synthesizing-Robust-Adversarial-Examples}
Anish Athalye, Logan Engstrom, Andrew Ilyas, and Kevin Kwok.
\newblock Synthesizing robust adversarial examples.
\newblock In \emph{Proceedings of the 35th International Conference on Machine
  Learning}, volume~80, pages 284--293, 2018.

\bibitem[Ben-Or et~al.(1988)Ben-Or, Goldwasser, and
  Wigderson]{ben2019completeness}
Michael Ben-Or, Shafi Goldwasser, and Avi Wigderson.
\newblock Completeness theorems for non-cryptographic fault-tolerant
  distributed computation.
\newblock In \emph{STOC}, pages 1--10, 1988.

\bibitem[Biggio et~al.(2013)Biggio, Corona, Maiorca, Nelson, {\v{S}}rndi{\'c},
  Laskov, Giacinto, and Roli]{Evasion-attacks}
Battista Biggio, Igino Corona, Davide Maiorca, Blaine Nelson, Nedim
  {\v{S}}rndi{\'c}, Pavel Laskov, Giorgio Giacinto, and Fabio Roli.
\newblock Evasion attacks against machine learning at test time.
\newblock In \emph{Joint European Conference on Machine Learning and Knowledge
  Discovery in Databases}, pages 387--402. Springer, 2013.

\bibitem[Blum et~al.(2005)Blum, Dwork, McSherry, and Nissim]{blum2005practical}
Avrim Blum, Cynthia Dwork, Frank McSherry, and Kobbi Nissim.
\newblock Practical privacy: the sulq framework.
\newblock In \emph{Proceedings of the twenty-fourth ACM SIGMOD-SIGACT-SIGART
  symposium on Principles of database systems}, pages 128--138, 2005.

\bibitem[Bogetoft et~al.(2009)Bogetoft, Christensen, Damg{\aa}rd, Geisler,
  Jakobsen, Kr{\o}igaard, Nielsen, Nielsen, Nielsen, Pagter,
  et~al.]{bogetoft2009secure}
Peter Bogetoft, Dan~Lund Christensen, Ivan Damg{\aa}rd, Martin Geisler, Thomas
  Jakobsen, Mikkel Kr{\o}igaard, Janus~Dam Nielsen, Jesper~Buus Nielsen, Kurt
  Nielsen, Jakob Pagter, et~al.
\newblock Secure multiparty computation goes live.
\newblock In \emph{International Conference on Financial Cryptography and Data
  Security}, pages 325--343. Springer, 2009.

\bibitem[Brendel and Bethge(2019)]{brendel2019approximating}
Wieland Brendel and Matthias Bethge.
\newblock Approximating cnns with bag-of-local-features models works
  surprisingly well on imagenet.
\newblock \emph{arXiv preprint arXiv:1904.00760}, 2019.

\bibitem[Brown et~al.(2017)Brown, Man{\'{e}}, Roy, Abadi, and
  Gilmer]{adversarialpatch}
Tom~B. Brown, Dandelion Man{\'{e}}, Aurko Roy, Mart{\'{\i}}n Abadi, and Justin
  Gilmer.
\newblock Adversarial patch.
\newblock \emph{CoRR}, abs/1712.09665, 2017.

\bibitem[Bubeck et~al.(2018)Bubeck, Price, and
  Razenshteyn]{bubeck2018adversarial}
S{\'e}bastien Bubeck, Eric Price, and Ilya Razenshteyn.
\newblock Adversarial examples from computational constraints.
\newblock \emph{arXiv preprint arXiv:1805.10204}, 2018.

\bibitem[{Carlini} and {Wagner}(2017)]{7958570}
N.~{Carlini} and D.~{Wagner}.
\newblock Towards evaluating the robustness of neural networks.
\newblock In \emph{2017 IEEE Symposium on Security and Privacy (SP)}, pages
  39--57, May 2017.
\newblock \doi{10.1109/SP.2017.49}.

\bibitem[Chaum et~al.(1988)Chaum, Cr{\'e}peau, and
  Damgard]{chaum1988multiparty}
David Chaum, Claude Cr{\'e}peau, and Ivan Damgard.
\newblock Multiparty unconditionally secure protocols.
\newblock In \emph{Proceedings of the twentieth annual ACM symposium on Theory
  of computing}, pages 11--19, 1988.

\bibitem[Cohen et~al.(2019)Cohen, Rosenfeld, and Kolter]{cohen2019certified}
Jeremy~M Cohen, Elan Rosenfeld, and J~Zico Kolter.
\newblock Certified adversarial robustness via randomized smoothing.
\newblock \emph{arXiv preprint arXiv:1902.02918}, 2019.

\bibitem[Dong et~al.(2018)Dong, Liao, Pang, Su, Zhu, Hu, and
  Li]{Dong_2018_CVPR}
Yinpeng Dong, Fangzhou Liao, Tianyu Pang, Hang Su, Jun Zhu, Xiaolin Hu, and
  Jianguo Li.
\newblock Boosting adversarial attacks with momentum.
\newblock In \emph{The IEEE Conference on Computer Vision and Pattern
  Recognition (CVPR)}, June 2018.

\bibitem[Dwork(2008)]{dwork2008differential}
Cynthia Dwork.
\newblock Differential privacy: A survey of results.
\newblock In \emph{International conference on theory and applications of
  models of computation}, pages 1--19. Springer, 2008.

\bibitem[Dwork and Nissim(2004)]{dwork2004privacy}
Cynthia Dwork and Kobbi Nissim.
\newblock Privacy-preserving datamining on vertically partitioned databases.
\newblock In \emph{Annual International Cryptology Conference}, pages 528--544.
  Springer, 2004.

\bibitem[Dwork et~al.(2006)Dwork, McSherry, Nissim, and
  Smith]{dwork2006calibrating}
Cynthia Dwork, Frank McSherry, Kobbi Nissim, and Adam Smith.
\newblock Calibrating noise to sensitivity in private data analysis.
\newblock In \emph{Theory of cryptography conference}, pages 265--284.
  Springer, 2006.

\bibitem[Engstrom et~al.(2017)Engstrom, Tsipras, Schmidt, and
  Madry]{Rotation_Translation}
Logan Engstrom, Dimitris Tsipras, Ludwig Schmidt, and Aleksander Madry.
\newblock A rotation and a translation suffice: Fooling cnns with simple
  transformations.
\newblock \emph{CoRR}, abs/1712.02779, 2017.

\bibitem[Evtimov et~al.(2017)Evtimov, Eykholt, Fernandes, Kohno, Li, Prakash,
  Rahmati, and Song]{evtimov2017robust}
Ivan Evtimov, Kevin Eykholt, Earlence Fernandes, Tadayoshi Kohno, Bo~Li, Atul
  Prakash, Amir Rahmati, and Dawn Song.
\newblock Robust physical-world attacks on deep learning models.
\newblock \emph{arXiv preprint arXiv:1707.08945}, 2017.

\bibitem[Fawzi et~al.(2018)Fawzi, Fawzi, and
  Fawzi]{DBLP:journals/corr/abs-1802-08686}
Alhussein Fawzi, Hamza Fawzi, and Omar Fawzi.
\newblock Adversarial vulnerability for any classifier.
\newblock \emph{CoRR}, abs/1802.08686, 2018.

\bibitem[Ford et~al.(2019)Ford, Gilmer, and Cubuk]{ford2019adversarial}
Nicolas Ford, Justin Gilmer, and Ekin~D. Cubuk.
\newblock Adversarial examples are a natural consequence of test error in
  noise, 2019.

\bibitem[Gilmer et~al.(2018)Gilmer, Metz, Faghri, Schoenholz, Raghu,
  Wattenberg, and Goodfellow]{DBLP:journals/corr/abs-1801-02774}
Justin Gilmer, Luke Metz, Fartash Faghri, Samuel~S. Schoenholz, Maithra Raghu,
  Martin Wattenberg, and Ian~J. Goodfellow.
\newblock Adversarial spheres.
\newblock \emph{CoRR}, abs/1801.02774, 2018.

\bibitem[Goldreich et~al.(1987)Goldreich, Micali, and
  Wigderson]{goldreich2019play}
Oded Goldreich, Silvio Micali, and Avi Wigderson.
\newblock How to play any mental game, or a completeness theorem for protocols
  with honest majority.
\newblock In \emph{STOC}, pages 218--229, 1987.

\bibitem[Goodfellow et~al.(2015)Goodfellow, Shlens, and Szegedy]{43405}
Ian Goodfellow, Jonathon Shlens, and Christian Szegedy.
\newblock Explaining and harnessing adversarial examples.
\newblock In \emph{International Conference on Learning Representations}, 2015.

\bibitem[Gu and Rigazio(2014)]{gu2014towards}
Shixiang Gu and Luca Rigazio.
\newblock Towards deep neural network architectures robust to adversarial
  examples.
\newblock \emph{arXiv preprint arXiv:1412.5068}, 2014.

\bibitem[He et~al.(2016)He, Zhang, Ren, and Sun]{DBLP:journals/corr/HeZR016}
Kaiming He, Xiangyu Zhang, Shaoqing Ren, and Jian Sun.
\newblock Identity mappings in deep residual networks.
\newblock \emph{CoRR}, abs/1603.05027, 2016.

\bibitem[Huang et~al.(2019)Huang, Liu, Pleiss, Van Der~Maaten, and
  Weinberger]{huang2019convolutional}
Gao Huang, Zhuang Liu, Geoff Pleiss, Laurens Van Der~Maaten, and Kilian
  Weinberger.
\newblock Convolutional networks with dense connectivity.
\newblock \emph{IEEE Transactions on Pattern Analysis and Machine
  Intelligence}, 2019.

\bibitem[Ilyas et~al.(2019)Ilyas, Santurkar, Tsipras, Engstrom, Tran, and
  Madry]{not-bugs}
Andrew Ilyas, Shibani Santurkar, Dimitris Tsipras, Logan Engstrom, Brandon
  Tran, and Aleksander Madry.
\newblock Adversarial examples are not bugs, they are features.
\newblock In \emph{Advances in Neural Information Processing Systems 32}, 2019.

\bibitem[Lee et~al.(2019)Lee, Yuan, Chang, and Jaakkola]{lee2019stratified}
Guang-He Lee, Yang Yuan, Shiyu Chang, and Tommi~S Jaakkola.
\newblock A stratified approach to robustness for randomly smoothed
  classifiers.
\newblock \emph{arXiv preprint arXiv:1906.04948}, 2019.

\bibitem[Li et~al.(2019{\natexlab{a}})Li, Qu, Li, Szurley, Kolter, and
  Metze]{li2019adversarial}
Juncheng Li, Shuhui Qu, Xinjian Li, Joseph Szurley, J~Zico Kolter, and Florian
  Metze.
\newblock Adversarial music: Real world audio adversary against wake-word
  detection system.
\newblock In \emph{Advances in Neural Information Processing Systems}, pages
  11908--11918, 2019{\natexlab{a}}.

\bibitem[Li et~al.(2019{\natexlab{b}})Li, Schmidt, and
  Kolter]{li2019adversarial2}
Juncheng Li, Frank Schmidt, and Zico Kolter.
\newblock Adversarial camera stickers: A physical camera-based attack on deep
  learning systems.
\newblock In \emph{International Conference on Machine Learning}, pages
  3896--3904, 2019{\natexlab{b}}.

\bibitem[Li(2019)]{bigballon2019cifarzoo}
Wei Li.
\newblock Cifar-zoo: Pytorch implementation of cnns for cifar dataset.
\newblock \url{https://github.com/BIGBALLON/CIFAR-ZOO}, 2019.

\bibitem[Madry et~al.(2017)Madry, Makelov, Schmidt, Tsipras, and
  Vladu]{madry2017towards}
Aleksander Madry, Aleksandar Makelov, Ludwig Schmidt, Dimitris Tsipras, and
  Adrian Vladu.
\newblock Towards deep learning models resistant to adversarial attacks.
\newblock \emph{arXiv preprint arXiv:1706.06083}, 2017.

\bibitem[Mahloujifar et~al.(2018)Mahloujifar, Diochnos, and
  Mahmoody]{DBLP:journals/corr/abs-1809-03063}
Saeed Mahloujifar, Dimitrios~I. Diochnos, and Mohammad Mahmoody.
\newblock The curse of concentration in robust learning: Evasion and poisoning
  attacks from concentration of measure.
\newblock \emph{CoRR}, abs/1809.03063, 2018.

\bibitem[Meng and Chen(2017)]{meng2017magnet}
Dongyu Meng and Hao Chen.
\newblock Magnet: a two-pronged defense against adversarial examples.
\newblock In \emph{Proceedings of the 2017 ACM SIGSAC Conference on Computer
  and Communications Security}, pages 135--147, 2017.

\bibitem[Moosavi-Dezfooli et~al.(2016)Moosavi-Dezfooli, Fawzi, and
  Frossard]{Moosavi-Dezfooli_2016_CVPR}
Seyed-Mohsen Moosavi-Dezfooli, Alhussein Fawzi, and Pascal Frossard.
\newblock Deepfool: A simple and accurate method to fool deep neural networks.
\newblock In \emph{The IEEE Conference on Computer Vision and Pattern
  Recognition (CVPR)}, June 2016.

\bibitem[Recht et~al.(2018)Recht, Roelofs, Schmidt, and
  Shankar]{DBLP:journals/corr/abs-1806-00451}
Benjamin Recht, Rebecca Roelofs, Ludwig Schmidt, and Vaishaal Shankar.
\newblock Do {CIFAR-10} classifiers generalize to cifar-10?
\newblock \emph{CoRR}, abs/1806.00451, 2018.

\bibitem[Samangouei et~al.(2018{\natexlab{a}})Samangouei, Kabkab, and
  Chellappa]{samangouei2018defense}
Pouya Samangouei, Maya Kabkab, and Rama Chellappa.
\newblock Defense-gan: Protecting classifiers against adversarial attacks using
  generative models.
\newblock \emph{arXiv preprint arXiv:1805.06605}, 2018{\natexlab{a}}.

\bibitem[Samangouei et~al.(2018{\natexlab{b}})Samangouei, Kabkab, and
  Chellappa]{samangouei2018defensegan}
Pouya Samangouei, Maya Kabkab, and Rama Chellappa.
\newblock Defense-{GAN}: Protecting classifiers against adversarial attacks
  using generative models.
\newblock In \emph{International Conference on Learning Representations},
  2018{\natexlab{b}}.

\bibitem[Sankaranarayanan et~al.(2018)Sankaranarayanan, Balaji, Castillo, and
  Chellappa]{Sankaranarayanan_2018_CVPR}
Swami Sankaranarayanan, Yogesh Balaji, Carlos~D. Castillo, and Rama Chellappa.
\newblock Generate to adapt: Aligning domains using generative adversarial
  networks.
\newblock In \emph{The IEEE Conference on Computer Vision and Pattern
  Recognition (CVPR)}, June 2018.

\bibitem[Santhanam and Grnarova(2018)]{santhanam2018defending}
Gokula~Krishnan Santhanam and Paulina Grnarova.
\newblock Defending against adversarial attacks by leveraging an entire gan.
\newblock \emph{arXiv preprint arXiv:1805.10652}, 2018.

\bibitem[Schott et~al.(2018)Schott, Rauber, Brendel, and
  Bethge]{DBLP:journals/corr/abs-1805-09190}
Lukas Schott, Jonas Rauber, Wieland Brendel, and Matthias Bethge.
\newblock Robust perception through analysis by synthesis.
\newblock \emph{CoRR}, abs/1805.09190, 2018.

\bibitem[Shafahi et~al.(2019)Shafahi, Huang, Studer, Feizi, and
  Goldstein]{shafahi2018are}
Ali Shafahi, W.~Ronny Huang, Christoph Studer, Soheil Feizi, and Tom Goldstein.
\newblock Are adversarial examples inevitable?
\newblock In \emph{International Conference on Learning Representations}, 2019.

\bibitem[Simonyan and Zisserman(2014)]{DBLP:journals/corr/SimonyanZ14a}
Karen Simonyan and Andrew Zisserman.
\newblock Very deep convolutional networks for large-scale image recognition.
\newblock \emph{CoRR}, abs/1409.1556, 2014.

\bibitem[Song et~al.(2017)Song, Kim, Nowozin, Ermon, and
  Kushman]{song2017pixeldefend}
Yang Song, Taesup Kim, Sebastian Nowozin, Stefano Ermon, and Nate Kushman.
\newblock Pixeldefend: Leveraging generative models to understand and defend
  against adversarial examples.
\newblock \emph{arXiv preprint arXiv:1710.10766}, 2017.

\bibitem[{Summers} and {Dinneen}(2019)]{8659168}
C.~{Summers} and M.~J. {Dinneen}.
\newblock Improved mixed-example data augmentation.
\newblock In \emph{2019 IEEE Winter Conference on Applications of Computer
  Vision (WACV)}, pages 1262--1270, Jan 2019.
\newblock \doi{10.1109/WACV.2019.00139}.

\bibitem[Szegedy et~al.(2014)Szegedy, Zaremba, Sutskever, Bruna, Erhan,
  Goodfellow, and Fergus]{Intriguing-properties-of-neural-networks}
Christian Szegedy, Wojciech Zaremba, Ilya Sutskever, Joan Bruna, Dumitru Erhan,
  Ian~J. Goodfellow, and Rob Fergus.
\newblock Intriguing properties of neural networks.
\newblock In \emph{2nd International Conference on Learning Representations},
  2014.

\bibitem[Tanay and Griffin(2016)]{DBLP:journals/corr/TanayG16}
Thomas Tanay and Lewis~D. Griffin.
\newblock A boundary tilting persepective on the phenomenon of adversarial
  examples.
\newblock \emph{CoRR}, abs/1608.07690, 2016.

\bibitem[Xie et~al.(2019)Xie, Wu, Maaten, Yuille, and He]{xie2019feature}
Cihang Xie, Yuxin Wu, Laurens van~der Maaten, Alan~L Yuille, and Kaiming He.
\newblock Feature denoising for improving adversarial robustness.
\newblock In \emph{Proceedings of the IEEE Conference on Computer Vision and
  Pattern Recognition}, pages 501--509, 2019.

\bibitem[Yao(1982)]{yao1982protocols}
Andrew~C Yao.
\newblock Protocols for secure computations.
\newblock In \emph{23rd annual symposium on foundations of computer science
  (sfcs 1982)}, pages 160--164. IEEE, 1982.

\bibitem[Yao(1986)]{yao1986generate}
Andrew Chi-Chih Yao.
\newblock How to generate and exchange secrets.
\newblock In \emph{27th Annual Symposium on Foundations of Computer Science
  (sfcs 1986)}, pages 162--167. IEEE, 1986.

\bibitem[Yun et~al.(2019)Yun, Han, Oh, Chun, Choe, and Yoo]{cutmix}
Sangdoo Yun, Dongyoon Han, Seong~Joon Oh, Sanghyuk Chun, Junsuk Choe, and
  Youngjoon Yoo.
\newblock Cutmix: Regularization strategy to train strong classifiers with
  localizable features.
\newblock \emph{CoRR}, abs/1905.04899, 2019.

\bibitem[Zhang et~al.(2017)Zhang, Ciss{\'{e}}, Dauphin, and Lopez{-}Paz]{mixup}
Hongyi Zhang, Moustapha Ciss{\'{e}}, Yann~N. Dauphin, and David Lopez{-}Paz.
\newblock mixup: Beyond empirical risk minimization.
\newblock \emph{CoRR}, abs/1710.09412, 2017.

\bibitem[Zheng et~al.(2016)Zheng, Song, Leung, and Goodfellow]{Zheng_2016_CVPR}
Stephan Zheng, Yang Song, Thomas Leung, and Ian Goodfellow.
\newblock Improving the robustness of deep neural networks via stability
  training.
\newblock In \emph{The IEEE Conference on Computer Vision and Pattern
  Recognition (CVPR)}, June 2016.

\end{thebibliography}
